\documentclass[sn-basic]{sn-jnl}

\usepackage{graphicx}%
\usepackage{multirow}%
\usepackage{amsmath,amssymb,amsfonts}%
\usepackage{amsthm}%
\usepackage{mathrsfs}%
\usepackage[title]{appendix}%
\usepackage{xcolor}%
\usepackage{textcomp}%
\usepackage{manyfoot}%
\usepackage{booktabs}%
\usepackage{algorithm}%
\usepackage{algorithmicx}%
\usepackage{algpseudocode}%
\usepackage{listings}%

\usepackage{geometry}
\usepackage{parskip}
\theoremstyle{thmstyleone}%
\theoremstyle{thmstyletwo}%

\theoremstyle{thmstylethree}%

\begin{document}

{
\setlength{\parskip}{0pt plus 1pt}  

\title[Article Title]{Using Instruction-Tuned Large Language Models to Identify Indicators of Vulnerability in Police Incident Narratives}

\author[1,3]{\fnm{Sam} \sur{Relins}}
\author[2,3]{\fnm{Daniel} \sur{Birks}}
\author*[1,3]{\fnm{Charlie} \sur{Lloyd}}

\affil[1]{\orgdiv{School for Business and Society}, \orgname{University of York}, \orgaddress{\city{York}, \country{UK}}}
\affil[2]{\orgdiv{School of Law}, \orgname{University of Leeds}, \orgaddress{\city{Leeds}, \country{UK}}}
\affil[3]{\orgdiv{ESRC Vulnerability and Policing Futures Research Centre}}
\affil*{\small This paper has been submitted to the Journal of Quantitative Criminology and is under review}

\abstract{
\textbf{Objectives}: Compare qualitative coding of instruction tuned large language models (IT-LLMs) against human coders in classifying the presence or absence of vulnerability in routinely collected unstructured text that describes police-public interactions. Evaluate potential bias in IT-LLM codings.\par\medskip

\textbf{Methods}: Analyzing publicly available text narratives of police-public interactions recorded by Boston Police Department, we provide humans and IT-LLMs with qualitative labelling codebooks and compare labels generated by both, seeking to identify situations associated with (i) mental ill health; (ii) substance misuse; (iii) alcohol dependence; and (iv) homelessness. We explore multiple prompting strategies and model sizes, and the variability of labels generated by repeated prompts. Additionally, to explore model bias, we utilize counterfactual methods to assess the impact of two protected characteristics - race and gender – on IT-LLM classification.\par\medskip

\textbf{Results}: Results demonstrate that IT-LLMs can effectively support human qualitative coding of police incident narratives. While there is some disagreement between LLM and human generated labels, IT-LLMs are highly effective at screening narratives where no vulnerabilities are present, potentially vastly reducing the requirement for human coding. Counterfactual analyses demonstrate that manipulations to both gender and race of individuals described in narratives have very limited effects on IT-LLM classifications beyond those expected by chance.\par\medskip

\textbf{Conclusions}: IT-LLMs offer effective means to augment human qualitative coding in a way that requires much lower levels of resource to analyze large unstructured datasets. Moreover, they encourage specificity in qualitative coding, promote transparency, and provide the opportunity for more standardized, replicable approaches to analyzing large free-text police data sources.
}

\maketitle
}

\newpage
\section{Introduction}\label{sec1}

The last decade has seen an increasing focus on the role of police in engaging with vulnerable populations. Police officers frequently come into contact with individuals experiencing mental health crises \citep{kane_police_2018, wood_what_2021}, homelessness \citep{herring_complaint-oriented_2019,kouyoumdjian_interactions_2019}, substance dependency \citep{winkelman_health_2018,zhang_relationship_2022} or exhibiting other complex needs. This shift has led to a reconceptualization of policing, moving away from the traditional ``warrior'' model focused solely on crime control, toward a ``guardian'' approach rooted in public protection, care, and community wellbeing \citep{engel_police_2015, wood_improving_2017}. In response, public health approaches to policing recognize that many societal challenges, such as mental health issues and addiction, require interventions that reach beyond traditional mechanisms of law enforcement, and instead advocate for multi-agency working, upstream preventive measures, and community-based support rather than punitive responses \citep{christmas_public_2019,van_dijk_law_2019}. Concurrently, trauma-informed practices in policing have also gained considerable traction in recent years, encouraging law enforcement to prioritize empathy, de-escalation, and compassionate communication, with the aim of reducing harm and enhancing community relations \citep{ko_creating_2008}.

Despite its policy relevance, quantifying the extent of police engagement with vulnerable populations remains challenging. Current quantitative estimates often rely on ``flags'' or ``markers'' — categorical data fields recorded in call and dispatch systems to indicate the presence of specific incident characteristics, which can include the presence of one or more predetermined types of vulnerability. The validity of such data as a means to measure police contact with vulnerable populations is questionable for a number of reasons. First, categorical indicators struggle to adequately capture the complexity of situations police often find themselves in, where the intersection of multiple vulnerabilities may blur the boundaries of predefined categories. Second, the identification of a given vulnerability depends on the judgement and discretion of police officers or call handlers, who may differ significantly in their assessments of specific situations. Third, markers may be inconsistently applied across incidents, potentially varying across individuals recording data and across incident types, creating disparities in how vulnerability is recognized, flagged and recorded.

These challenges are likely reflected in the significant variation observed in efforts to quantify police engagement with vulnerability. To illustrate, relying on several representative snapshots of police demand, the UK's Policing Productivity Review \citep{homeoffice_policing_2023} estimated that between 5\% and 9\% of incidents involve mental ill health. Conversely, evidence submitted to a UK Parliamentary Inquiry into Policing and Mental Health on behalf of all UK Police Forces estimated that 20\% of police time was spent dealing with mental health related calls \citep{home_affairs_select_committee_evidence_2015a}, and that over 40\% of calls for service were associated with those deemed vulnerable \citep{home_affairs_select_committee_evidence_2015b}. Similarly, a systematic review of 15 studies conducted in North America estimated that approximately 1\% of calls for service involved individuals with mental disorders \citep{livingston_contact_2016}. However, estimates varied substantially depending on the identification method used, ranging from 1\% in dispatcher coding to 6\% in police officer surveys and 9\% in fieldworker observations. Such disparities between measurement approaches underscore the limitations of routine data collection, and the broader challenge of defining and recognizing vulnerability in policing contexts.

One potentially rich source of information in this context lies in the unstructured text narratives that police officers or call center staff record during routine logging of incidents or calls-for-service. These narratives typically document the circumstances, behaviors, and contextual details surrounding an incident and are collected for a variety of operational reasons including providing context for responding officers, documenting events for evidentiary purposes, and ensuring accountability and oversight. Yet despite their potential to provide insights beyond standardized data fields, such narratives remain largely underutilized in efforts aimed at quantifying police involvement with vulnerability. The primary reason for this lies in the resource-intensive nature of traditional analytical methods capable of deriving insights from unstructured data, which demand significant manual effort and are often infeasible at scale. Ultimately, this may limit agencies’ ability to access detailed insights that could otherwise support evidence-based problem and demand analyses, training, and inter-agency coordination \citep{dixon_improving_2021}.

Recent advances in large language models (LLMs) offer new ways to automate the processing of unstructured text data. The latest instruction-tuned LLMs (IT-LLMs) are designed to interpret and respond to natural language instructions directly, enabling them to flexibly support complex tasks like qualitative coding without additional specialized training \citep{zhang_instruction_2024}. By enabling scalable, qualitative coding of free-text data, IT-LLMs may provide viable means to bridge the gap between the limited scope of structured data and the detailed but labor-intensive nature of narrative analysis. In this study, we assess the capacity of IT-LLMs to replicate a deductive coding exercise: using unstructured incident narratives from the Boston Police Department, we prompt LLMs to generate labels designed to identify situations associated with (i) mental ill health; (ii) substance misuse; (iii) alcohol dependence; and (iv) homelessness. We then compare the LLM-generated labels with those produced by non-expert human coders. Rather than seeking to definitively estimate the prevalence of vulnerabilities within this specific dataset, our primary aim is to explore the viability of a scalable methodology that could subsequently be deployed to generate such estimates.

The remainder of this paper is organized as follows: we first provide a brief overview of the development of instruction tuned large language models, followed by a discussion on their emerging applications within policing and qualitative analysis. We then describe our specific methodological approach, before presenting our primary findings on the alignment between LLM and human-coded labels and further analyses on potential model biases. In the concluding sections, we discuss the implications of findings for police research and practice, exploring the potential for LLMs to augment traditional qualitative methods and support more data-informed policing strategies.

\subsection{The Development of Instruction Tuned Large Language Models}\label{subsec11}

Over the past decade, the capabilities of generative language models - artificial intelligence systems designed to produce human-like text - have transformed dramatically. At their core, these models are trained to perform a simple task: predicting the next word in a sequence based on the preceding words. Given the phrase ``The sun is shining,'' for example, the model might suggest ``brightly'' or ``today'' as natural continuations, drawing on patterns of word usage in its training data. Early progress in applying deep learning to natural language tasks lagged behind other machine learning domains, such as computer vision. While Recurrent Neural Networks (RNNs) \citep{elman_finding_1990} and their variant Long Short-Term Memory (LSTM) networks \citep{hochreiter_long_1997} marked important early breakthroughs in language modelling, they faced significant limitations. These models could generate coherent text in small fragments but struggled with longer passages \citep{bengio_learning_1994}. Their sequential processing of inputs also made them computationally inefficient, limiting both model size and training capacity.

The introduction of the transformer architecture in 2017 marked a pivotal advancement in generative language models \citep{vaswani_attention_2023}. Transformers overcame the long-range dependency issue present in earlier models through the use of a self-attention mechanism, which allowed them to effectively capture relationships between distant words in a sequence. More importantly, transformers could process text in parallel rather than sequentially, enabling far greater computational efficiency and allowing models to scale up significantly in both size and training data. Early transformer-based models demonstrated remarkable improvements in formal language representation compared to RNNs and LSTMs, producing long, coherent texts with correct grammar and syntax. As these models grew in scale, they began exhibiting 'emergent capabilities' — performing a diverse range of tasks without any task-specific pre-training \citep{radford_language_2019, brown_language_2020}. These behaviors demonstrated the potential for language models to generalize across a wider range of applications than previously anticipated.

The current generation of instruction-tuned language models represents a further significant advance in this trajectory. These models are explicitly trained to interpret and follow natural language instructions, enabling them to adapt flexibly to diverse tasks while maintaining coherent reasoning \citep{mishra_cross-task_2022,wei_chain--thought_2023}. Models such as GPT, Claude, and Llama can now engage in sophisticated tasks including analysis, summarization, and complex problem-solving - activities that previously required human expertise \citep{kojima_large_2022,srivastava_beyond_2023}. This capability to follow explicit instructions while drawing on broad knowledge has transformed these models from simple text generators into versatile tools for knowledge work, opening new possibilities for automating complex cognitive tasks that were previously considered beyond the reach of computational approaches.

\subsection{Related Work}\label{subsec12}

Instruction-tuned large language models (IT-LLMs) are a recent innovation, and research applying them remains in its infancy. Most natural language processing work in policing has focused on rule-based systems and unsupervised methods for crime classification, entity extraction, and summarization \citep{ku_crime_2008, srihari_supporting_2008, elzinga_terrorist_2010, poelmans_formally_2011, kuang_crime_2017, guetterman_augmenting_2018, karystianis_automatic_2018, karystianis_automated_2019, karystianis_text_2024, johnsen_impact_2019, birks_unsupervised_2020, lwin_tun_supporting_2023}. While supervised learning and early transformer models like BERT have shown some promise \citep{haleem_automated_2019, osorio_enhancing_2020, langton_policing_2021, halford_anti-social_2022, barros_leveraging_2023, hodgkinson_domestic_2023}, their task-specific training inherently limits their flexibility compared to the adaptable nature of IT-LLMs. Although some papers discuss potential uses of IT-LLMs in police work \citep{dubravova_artificial_2024,adams_large_2024,puczynska_large_2024}, these remain largely theoretical, with no empirical studies to the best of our knowledge directly applying IT-LLMs for qualitative coding in policing contexts.

Outside of policing research, a growing body of work has investigated the potential of IT-LLMs specifically for deductive coding tasks common in qualitative research. Deductive coding is a structured approach where text is reviewed to identify and label instances of specific, predefined categories, allowing researchers to apply consistent criteria across data. A few studies have shown promising results, indicating that IT-LLMs can closely replicate human-generated labels, often achieving comparable levels of inter-rater reliability (IRR) \citep{xiao_supporting_2023,chew_llm-assisted_2023,ashwin_using_2023,tai_examination_2024,dunivin_scalable_2024}. This work explores various prompting techniques, including zero-shot, few-shot, and chain-of-thought reasoning \citep{xiao_supporting_2023}, as well as different sizes of language model, with results suggesting that larger models and sophisticated prompting approaches can enhance coding accuracy \citep{dunivin_scalable_2024}. Some research has also examined critical aspects like output consistency and potential bias in LLM-based coding: \cite{tai_examination_2024} explore the variability of LLM responses across repeated prompts, while \cite{ashwin_using_2023} analyzed biases tied to demographic attributes. Together, these studies indicate the potential of applying IT-LLMs in deductive coding, and the need for careful evaluation to determine the effects of variability and bias.

As far as we are aware, no existing studies have applied IT-LLMs in a policing context to empirically assess their capacity for deductive coding tasks. Our study addresses this gap, specifically exploring IT-LLMs' potential to identify vulnerabilities within unstructured police narratives. Moreover, we are the first to investigate the comparative effects of model size, prompting strategies and how these variables impact labelling accuracy and consistency across repeated prompts. Additionally, we perform counterfactual analyses to assess potential biases in LLM outputs, systematically testing for demographic influences by modifying attributes within coding tasks. By uniting these dimensions in a single study, we provide new insights into both the application of IT-LLMs within policing research and the methodological considerations necessary for applying LLMs to qualitative coding more broadly.

\section{Our Approach}\label{sec2}

\subsection{Dataset}\label{subsec21}

We evaluated IT-LLMs' effectiveness in deductive coding using narrative data from the Boston Police Department's field interrogation and observation (FIO) dataset \citep{analyze_boston_bpd_nodate}. These narratives consist of free-text descriptions that document police interactions with the public, including sufficient contextual detail to identify vulnerabilities such as homelessness and substance abuse when present. The data are released under an Open Data Commons Public Domain Dedication and License (PDDL)\footnote{Full text of the PDDL can be found at: http://opendatacommons.org/licenses/pddl/1.0/}, permitting both their use with commercial LLM services and enabling other researchers to independently replicate our analysis.

Two example narratives are included below - note the use of redaction to remove person-specific identifiers, and the use of domain specific shorthand:

\begin{quote}
Example 1:

“xxx has been seen walking on dorchester ave and hanging in fields corner. xxx spoke with officers and stated that she has a drinking problem and is homeless and hangs in the fields corner area. h983 sgt det cullity to be notified. very minor bop. hk01f - fritch/moccia”
\end{quote}

\begin{quote}
Example 2:

“officers observed xxx in the area, approaching multiple pedestrians, in the street, and on the sidewalk. xxx was observed constantly walking back and forth on the street, on dorchester ave. officers conducted a threshold inquiry, xxx stated he was looking for directions to jfk/red line, then recanted and said he was looking to meet a girl to possibly have drinks, and also said that he is in aa. he lives in hingham. distribution of class b on record. for intel fio - taylor/moccia h425f”
\end{quote}

To prepare the dataset for analysis, we extracted the narrative texts from all records where narratives were available, those recorded between June 2015 and December 2023. We cleaned the data by adding spaces after punctuation and the redacted content (“XXX”) and removing unnecessary whitespace. Duplicate records were eliminated, and any narratives with fewer than 200 characters were excluded from the dataset. This process resulted in a final dataset comprising 32,218 unique narrative texts with a median word count of 81.

\subsection{Codebook Development}\label{subsec22}

We developed a codebook focusing on four specific vulnerabilities: mental ill health, substance misuse, alcohol dependence, and homelessness. These vulnerabilities were selected as anecdotal evidence suggested they feature in a significant number of police interactions, they could be relatively easily defined and were recognizable concepts to non-experts. To develop the codebook, we manually selected 100 narratives from those used in prompt development (discussed in section 2.3 below) detailing a range of cases relating to the selected vulnerabilities, from clear examples to those with indirect or circumstantial elements, providing a basis for determining the threshold of evidence required to identify each vulnerability. Each member of the research team independently coded these narratives based on their intuitive understanding. We then compared and discussed these initial codes to reach consensus definitions for each vulnerability which formed our codebook. The final definitions included detailed definitions and examples, designed such that non-expert audiences, such as call-center administrative staff, could apply them without needing further input.

Recognizing that many examples contained ambiguous evidence, we adopted a three-tiered labelling scheme: positive, inconclusive, and negative. This approach allowed us to better capture the uncertainty inherent in many narratives, where vulnerability indicators were often implicit rather than explicit (e.g., circumstantial cues rather than direct statements). By including an inconclusive category, we aimed to accommodate this ambiguity without forcing definitive positive or negative labels on cases lacking clear evidence.

Appendix A contains codebook definitions of all vulnerabilities considered. 

\subsection{Prompt Development}\label{subsec23}

The language models were provided with text instructions, known as prompts, that describe the deductive coding task. The design and choice of prompts are important, as they affect the model’s behavior and the quality of its outputs, as shown in numerous recent studies exploring the performance of IT-LLMs \citep{white_prompt_2023}. In this study, we tested two prompting approaches: a basic codebook prompt that used the codebook definitions verbatim and a custom prompt iteratively refined to optimize the model’s performance. Note that the narrative data used in the development and testing of prompts were distinct from that used in all subsequent experiments.

\subsubsection{Codebook Prompt}\label{subsubsec231}

The codebook prompt quoted the definitions of vulnerabilities from the codebook verbatim. This approach was designed to directly compare the model’s ability to interpret the same information that would be provided to human coders, and to investigate whether definitions designed for humans are sufficient for language models to follow accurately. Additionally, this method closely mirrors the traditional manual coding process, differing primarily in substituting the human coder with an LLM. By simply providing the codebook definitions alongside a short set of task instructions, it introduces minimal additional complexity, preserving the familiar workflow while enabling the use of LLMs for coding tasks.

The codebook prompt is a minimal template based on the codebook definitions. It instructs the model to read the codebook definition and then classify police narratives using that definition. The prompt instructs the model on the desired response: brief notes highlighting relevant quotes from the narrative and linking them to the respective parts of the codebook definition. This approach was informed by the “chain of thought” method \citep{wei_chain--thought_2023}, where the language model generates intermediate reasoning steps leading to a final answer.  Subsequently, the model is instructed to generate a classification in a pre-defined format that can be parsed by a processing script. If the model’s response fails to be parsed correctly, it is sent an additional message asking for reformatting. This process is repeated up to three times, after which, if the response is still incorrect it is marked as empty/missing.

To maintain a valid comparison between the LLM’s interpretation of codebook definitions and human coding, we deliberately kept the template instructions for the model simple and refrained from experimenting with them to improve classifications.

The codebook prompt template is as follows:

\begin{verbatim}
    Read the following definition:

    {{ vulnerability definition }}

    You will be provided police incident reports, and should use the 
    definition to classify the report.

    Your response should begin with short notes highlighting quotes from 
    the report, and aligning them with quotes from the definitions above. 
    Keep your notes brief, 2 sentences max. Follow the highlighted evidence 
    with a classification that aligns with the evidence and the definitions. 
    Return your classification in the following format: 

    `Classification: [POSITIVE, INCONCLUSIVE, NEGATIVE]`. 

    Ensure that your response ends with your classification or it will be 
    rejected.
\end{verbatim}

And the parsing failure prompt:

\begin{verbatim}
    "The response you've provided does not conform to the format 
    requested. Please classify the log in the following format:

    Classification: [POSITIVE, INCONCLUSIVE, NEGATIVE]"
\end{verbatim}

\subsubsection{Custom Prompt}\label{subsubsec232}

In contrast to the codebook prompt, we also developed custom prompts to optimize the instructions specifically for the LLMs.. This approach, known as “prompt engineering,” involves iteratively testing and improving the instructions given to a language model to achieve the desired model behavior. Initially, the custom prompt began as a minimal set of instructions, asking the model to identify instances of a given vulnerability, without providing a specific definition of that vulnerability, and to label them as positive, inconclusive, or negative based on the evidence present in the narrative. This approach relies on the model’s ‘understanding’ of concepts such as mental ill health as encoded through its training data. This baseline prompt was then refined by analyzing the LLM’s outputs and adding or rephrasing instructions to address any errors or biases observed in its responses. This process allowed us to fine-tune the instructions to better suit the LLM’s strengths and limitations and reduce the size and complexity of instructions with comparison to the codebook prompts, which may be advantageous especially for smaller models. 

The final custom prompt template reflects this process of iterative refinement. The template retains the core task description and labelling scheme from the initial minimal prompt but includes more detailed and specific instructions. We added phrases like “contains unmistakable evidence of, having ruled out any other plausible explanations” and “evidence that is best explained by… but there is not definitive or conclusive confirmation…” based on initial experiments showing the model was too permissive in its assignment of positive and inconclusive labels. Additional vulnerability-specific instructions were added to the template to address any specific mistakes and steer toward the desired classifications for each vulnerability.

The custom prompt template reads as follows:

\begin{verbatim}
    You are required to classify police incident reports for involvement 
    of persons experiencing {{ vulnerability }}. Use the following 
    definitions for the labels you should assign:

    POSITIVE: Report confirms that someone is experiencing 
    {{ vulnerability }}, or contains unmistakable evidence of 
    {{ vulnerability }} having ruled out any other plausible 
    explanations. For example:

    {{ positive_evidence }}

    INCONCLUSIVE: Report contains evidence that is best explained by 
    an individual experiencing {{ vulnerability }}, but there is not 
    definitive or conclusive confirmation of {{ vulnerability }}. 
    For example:

    {{ inconclusive_evidence }}

    NEGATIVE: Evidence for {{ vulnerability }} that can be explained 
    by other factors, or no evidence for {{ vulnerability }}. The 
    following should not be considered evidence for {{ vulnerability }}:

    {{ negative_evidence }}

    Write short notes highlighting quotes from the report, and link 
    each quote to the relevant quote above. Keep the notes to two 
    sentences max.

    End your report with a classification that aligns with the 
    evidence you have highlighted. Use the format “Classification: 
    [POSITIVE, INCONCLUSIVE, NEGATIVE]”. If the final word of your 
    report is not the classification, it will be marked invalid.
\end{verbatim}

The following is an example of the specific instructions for one of the vulnerabilities, in this case “homelessness”:

\begin{verbatim}
    Positive Evidence:
      - Statements that someone is homeless or does not have any 
      nighttime accommodation
      - Individuals engaging with or being offered homelessness 
      services or organizations
      - Individuals being found with makeshift sleeping/living 
      arrangements on the street or in unstable living environments

    Inconclusive Evidence:
      - Strong evidence that someone does not have any nighttime 
      accommodation, but is not definitive
      - Being found asleep in public

    Negative Evidence:
      - Causing a nuisance, loitering, or being trespassed from places 
      where homeless individuals may congregate
      - Use of detox or rehab services for alcohol or drug abuse
      - Any evidence or behavior associated with a person's drug use, 
      alcoholism, mental health difficulties, or sex work
      - Any vague or uncooperative responses to police questioning 
      about address information that don't result in an admission of 
      homelessness
\end{verbatim}

Appendix B contains custom prompts for all vulnerabilities considered. 

\subsection{LLMs}\label{subsec24}

In addition to testing different prompting strategies, we evaluated different IT-LLMs to assess their capabilities for coding tasks. Recent advancements have produced a range of models differing in size, measured by the number of parameters, and whether they are open-source or proprietary. These factors significantly influence their performance, cost, and suitability for deployment in circumstances where computational resources are scarce or where data security considerations limit the sharing of data with third parties.

LLMs range in size from small models with around 1 billion parameters to extremely large models exceeding 500 billion parameters. Larger models tend to excel in handling complex and verbose instructions, performing better in tasks requiring nuanced understanding and reasoning skills. However, this comes at the cost of requiring advanced and expensive hardware, typically accessed through cloud-computing services. Smaller models, though generally less capable in handling complex tasks, are far more computationally efficient. They can run on consumer-grade hardware, such as laptops or smartphones, making them both cost-effective and widely accessible. This efficiency makes them ideal for scenarios with limited computational resources or strict data security requirements that preclude the use of cloud-based services. Balancing these trade-offs—between performance, cost, and deployment constraints—is critical when selecting a model, especially when working with sensitive data, such as police narratives, that cannot be shared with third parties.

Another important distinction is whether models are open-source or proprietary. Open-source models are freely available, with permissive licenses that allow unrestricted use, modification, and deployment on private hardware. This flexibility is particularly valuable for sensitive data, as it enables full control over the environment and ensures compliance with governance requirements. Open-source models also encourage transparency, allowing methods to be easily replicated, scrutinized, and improved upon by other researchers. Proprietary models, developed by companies like OpenAI and Google, often represent the state of the art in performance but are accessed through paid cloud services which limit how the models are used or modified, and require sharing data with the provider’s infrastructure. Furthermore, their high costs and the potential for changing usage terms add additional considerations for long-term projects. 

For our study, we tested models of varying sizes and both open-source and proprietary nature to evaluate their performance in analyzing police incident narratives. Specifically, we used the following models:
\begin{enumerate}
    \item \textbf{Llama 8B \& 70B}: These open-source models, released by Meta, have shown competitive performance relative to their size. The 8 billion parameter model represents a smaller, more accessible option, while the 70 billion parameter model provides a mid-sized alternative with enhanced capabilities.
    \item \textbf{GPT-4o}: This proprietary model from OpenAI is rumored to have over 1 trillion parameters, representing the state-of-the-art in LLM performance. While the exact size is undisclosed, GPT-4o is known for its advanced capabilities and state-of-the-art performance across a wide range of tasks.
\end{enumerate}
Hereon, we shall use the term ``labelling-configuration'' to describe each unique combination of a model and a prompting strategy used to classify narratives. In total, we tested five configurations: Codebook and Custom prompts with Llama 8B and 70B, and the Codebook prompt with GPT-4o. For example, the configuration “Codebook 70B” refers to using the codebook prompt with the Llama 70B model.

\subsection{Label Variability}\label{subsec25}

IT-LLMs generate text probabilistically, meaning they may produce different outputs given identical inputs. For our coding task, this means a single narrative could receive different vulnerability labels across multiple classifications, even when using the same model and prompt. To assess this potential variability in coding decisions, we classified each narrative ten times for each vulnerability using identical labelling configurations and analyzed the consistency of these repeated classifications.

\subsection{Evaluation Dataset}\label{subsec26}

We began by using Llama-based models (7B and 80B variants with both custom and codebook prompts) to code 4,000 randomly selected narratives\footnote{The quantity of narratives and choice of Llama models for initial coding were solely determined by the cost implications of using cloud-computing services from which models were accessed}. This approach helped estimate the level of class-imbalance within our dataset - recognizing that certain vulnerabilities were unlikely to be prevalent across all narratives - and in turn directed a purposive sampling of a subset of narratives for evaluation. For each narrative, we generated ten labels per labelling configuration to capture the inherent variability in LLM outputs. We then implemented a consensus approach: each narrative's final label was determined by the majority across its ten generated labels. Where no majority emerged, the narrative was marked as inconclusive.

Results of these analyses, shown in Table 1, indicate that the majority of police narratives were classified as not containing evidence of the specified vulnerabilities. Notably, the custom prompt configurations produced the highest proportions of negative classifications, often exceeding 90\%, while the codebook prompts resulted in more variability, particularly with the 8B model. Table 2 depicts a significant disparity in the unanimity of negative labels, with the custom prompts configurations consistently achieving higher rates of unanimous negative classifications compared to the codebook prompt configurations.

\begin{table}[h]
\begin{tabular}{|l|l|r|r|r|r|}
    \hline
    \textbf{Vulnerability} & \textbf{Label} & \textbf{Custom 8B} & \textbf{Custom 70B} & \textbf{Codeb.k 8B} & \textbf{Codeb.k 70B} \\
    \hline
    \multirow{3}{*}{\textbf{Alcohol Dep.}} & \textbf{Negative} & 94.2 & 96.150 & 45.975 & 89.950 \\
     & \textbf{Inconclusive} & 5.275 & 3.475 & 27.5 & 8.325 \\
     & Positive & 0.525 & 0.375 & 26.525 & 1.725 \\
    \hline
    \multirow{3}{*}{\textbf{Substance Misuse}} & \textbf{Negative} & 87.9 & 87.350 & 42.925 & 74.7 \\
     & \textbf{Inconclusive} & 9.225 & 8.625 & 30.675 & 17.125 \\
     & \textbf{Positive} & 2.875 & 4.025 & 26.4 & 8.175 \\
    \hline
    \multirow{3}{*}{\textbf{Homelessness}} & \textbf{Negative} & 93.725 & 88.925 & 51.750 & 74.050 \\
     & \textbf{Inconclusive} & 3.275 & 7.350 & 36.725 & 21.925 \\
     & \textbf{Positive} & 3.0 & 3.725 & 11.525 & 4.025 \\
    \hline
    \multirow{3}{*}{\textbf{Mental Ill Health}} & \textbf{Negative} & 93.2 & 95.8 & 47.5 & 81.725 \\
     & \textbf{Inconclusive} & 5.225 & 2.875 & 26.425 & 14.175 \\
     & \textbf{Positive} & 1.575 & 1.325 & 26.075 & 4.1 \\
    \hline
\end{tabular}
\caption{Distribution of consensus labels across different labelling configurations and vulnerabilities. Values show the percentage of narratives assigned each label type (negative, inconclusive, positive) based on majority voting across 10 iterations. Results demonstrate that custom prompts generally produced more negative classifications than codebook prompts, with the effect particularly pronounced for smaller models.}
\label{tab:label-distribution}
\end{table}
\begin{table}[h]
\begin{tabular}{|l|r|r|r|r|}
    \hline
    \textbf{Vulnerability} & \textbf{Codebook 70B} & \textbf{Codebook 8B} & \textbf{Custom 70B} & \textbf{Custom 8B} \\
    \hline
    \textbf{Alcohol Dependence} & 84.150 & 12.000 & 94.150 & 90.525 \\
    \hline
    \textbf{Substance Misuse} & 65.050 & 13.050 & 82.775 & 80.750 \\
    \hline
    \textbf{Homelessness} & 57.150 & 14.225 & 83.250 & 90.775 \\
    \hline
    \textbf{Mental Ill Health} & 70.400 & 15.000 & 92.825 & 88.100 \\
    \hline
\end{tabular}
\caption{Proportion of narratives receiving unanimous negative labels (10/10 votes) across different labelling configurations and vulnerabilities. Results show that custom prompts achieved consistently higher rates of unanimous negative classifications compared to codebook prompts, with larger models generally producing more unanimous classifications than smaller ones.}
\label{tab:unanimous-labels}
\end{table}

Informed by these analyses we designed a sampling method to include a higher proportion of positive and inconclusive labels for each vulnerability when selecting an evaluation subset of 500 narratives to be coded by humans\footnote{Again, the selection of 500 narratives was simply constrained by resources available}. To ensure a balanced evaluation, we used the consensus labels from the custom 70B configuration as our reference, based on our preliminary experimentation, that suggested larger models with custom prompts provide the most accurate labels. We randomly selected 100 examples labelled as negative for each of the four vulnerabilities. We then randomly selected 50 examples labelled as positive and 50 labelled as inconclusive for each of the four vulnerabilities - given that several examples had non-negative labels for multiple vulnerabilities, the final label proportions were actually greater than 50. For alcohol dependence, however, only 15 examples were marked positive, so we supplemented this with 35 inconclusive examples to maintain sample size.

Due to the high API costs associated with using GPT-4o, we limited our coding of the police narratives to the evaluation dataset of 500 narratives, as opposed to the 4000 narratives coded by the Llama configurations. We also chose to limit the evaluation of GPT-4o to the codebook prompt only, without developing a custom prompt. The decision to avoid a custom prompt for GPT-4o was based on the substantial API costs of prompt refinement, and that our initial experiments didn’t suggest that there would be much value in testing a custom prompt. We proposed that the GPT-4o model (the largest utilized) would be best suited to the detailed and lengthy instructions in the codebook, and that custom prompts would be most valuable to smaller models, where more carefully worded instructions can yield more dramatic improvements in outputs.

To code the 500 examples, we recruited two non-expert human coders who had not previously conducted qualitative coding of incident narratives and were not experts on any of the specified vulnerabilities. Each coder was given basic instructions to code each narrative for the four vulnerabilities using the definitions provided in the codebook. Coders worked independently without conferring and were instructed to use their own intuition whenever the guidance in the codebook was unclear. This approach aimed to reflect the codes that might be applied by police administrative staff, rather than an academic coding exercise. Any examples where the two coders disagreed were subsequently reviewed by the research team, who adjudicated disagreements to assign a consensus label, resulting in a single set of human labels for comparison with the LLM outputs. The numbers (and percentages) of examples requiring adjudication were as follows: mental ill health 49 (9.8\%); substance misuse 118 (23.6\%); alcohol dependence 55 (11\%); homelessness 50 (10\%).

\section{Analysis \& Results}\label{sec3}

\subsection{LLM Consensus vs Humans}\label{subsec31}

The following analyses compare LLM consensus labels from each labelling-configuration with those generated by human coders. As discussed previously, the LLM consensus label was determined by majority vote among ten labels generated for each narrative. If no majority label was present, the label was marked as inconclusive. 

\subsubsection{Error Analysis}\label{subsubsec311}

To quantify disagreement between the LLM consensus and the human labels, we assigned numerical values to each label: negative (0), inconclusive (1), and positive (2). We then calculated the mean squared error (MSE) for each set of LLM labels compared to the human labels. The MSE is given by Eq. (1):

\begin{equation}
    MSE = \frac{1}{n}\sum_{i=1}^{n} (y_i - \hat{y_i})^2
\end{equation}

where $y_i$ represents the human label for example $i$, $\hat{y}_i$ represents the LLM consensus for the same example, and $n$ is the number of examples. Scoring labels as 0, 1, and 2 allows MSE to capture the severity of disagreements, penalising larger mismatches (e.g., negative vs positive) more than smaller ones (e.g., negative vs inconclusive)

\begin{figure}[htbp]
    \centering
    \includegraphics[]{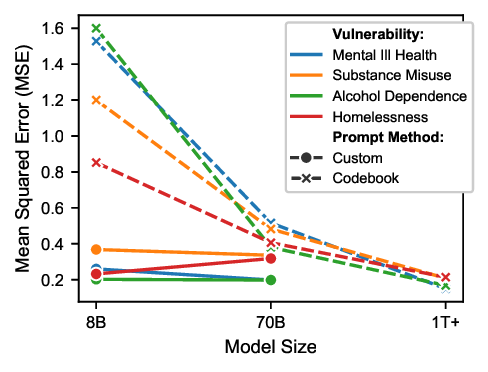}
    \caption{Mean squared error (MSE) between human and LLM consensus labels across different model sizes (8B, 70B, and 1T+) and prompt methods (Custom and Codebook) for four vulnerability types. Solid lines with circles represent Custom prompts, while dashed lines with crosses represent Codebook prompts.}
    \label{fig:mse-plot}
\end{figure}

Fig. \ref{fig:mse-plot} displays the MSE results for each LLM size and prompting method. A clear trend is observed in the codebook configurations, with errors decreasing as model size increases. Errors decrease significantly from the 8B model to the 70B model, with a smaller decline for the larger GPT-4o model. This trend demonstrates a general improvement in alignment with human labels for the codebook prompting approach as model size increases. The custom prompt configurations exhibit considerably lower errors compared to their codebook counterparts and are comparable in performance to the much larger GPT-4o codebook configuration. However, unlike the codebook configurations, there is no obvious trend in MSE reduction as model size increases for the custom configurations. The errors fall slightly for mental ill health, remain largely the same for substance misuse and alcohol dependence, and show a moderate increase for homelessness.

\subsubsection{Precision, Recall, \& F1 Score}\label{subsubsec312}

To gain a more precise understanding of the performance of various labelling configurations, with particular focus on the less frequent positive and inconclusive labels, we calculated precision, recall and F1 scores. To do so, we grouped the inconclusive and positive labels into a combined ``positive'' category, treating the negative labels as a separate category. We then calculated the statistics as follows:

\begin{itemize}
    \item \textbf{Precision:} Precision measures the accuracy of the positive labels assigned by the models. It is defined as the proportion of true positive labels among the positive labels the model assigned, shown in Eq.~\eqref{eq:precision}:
    
    \begin{equation}
        Precision = \frac{True\;Positives}{True\;Positives + False\;Positives}
        \label{eq:precision}
    \end{equation}
    \vspace{3pt}
    
    A higher precision indicates that the model makes fewer false positive errors, meaning that the positive identifications are more likely to be correct.

    \item \textbf{Recall:} Recall, also known as sensitivity, assesses the model's ability to identify all actual positive cases. It is defined in Eq.~\eqref{eq:recall}, as the proportion of true positive cases among all actual positive cases:
    
    \begin{equation}
        Recall = \frac{True\;Positives}{True\;Positives + False\;Negatives}
        \label{eq:recall}
    \end{equation}
    \vspace{3pt}
    
    A higher recall indicates that the model is more effective at detecting positive cases, reducing the number of false negatives.

    \item \textbf{F1 Score:} The F1 score provides a balance between precision and recall, offering a single metric that accounts for both false positives and false negatives, shown in Eq.~\eqref{eq:f1}:
    
    \begin{equation}
        F1\;Score = 2 \times \frac{Precision \times Recall}{Precision + Recall}
        \label{eq:f1}
    \end{equation}
    \vspace{3pt}
    
    F1 is particularly useful when dealing with imbalanced datasets, as it harmonizes the need for both high precision and high recall. A higher F1 score indicates a better overall performance of the model in classifying the positive labels correctly while minimizing both types of errors.
\end{itemize}

\begin{figure}[htbp]
    \centering
    \includegraphics[width=0.98\textwidth]{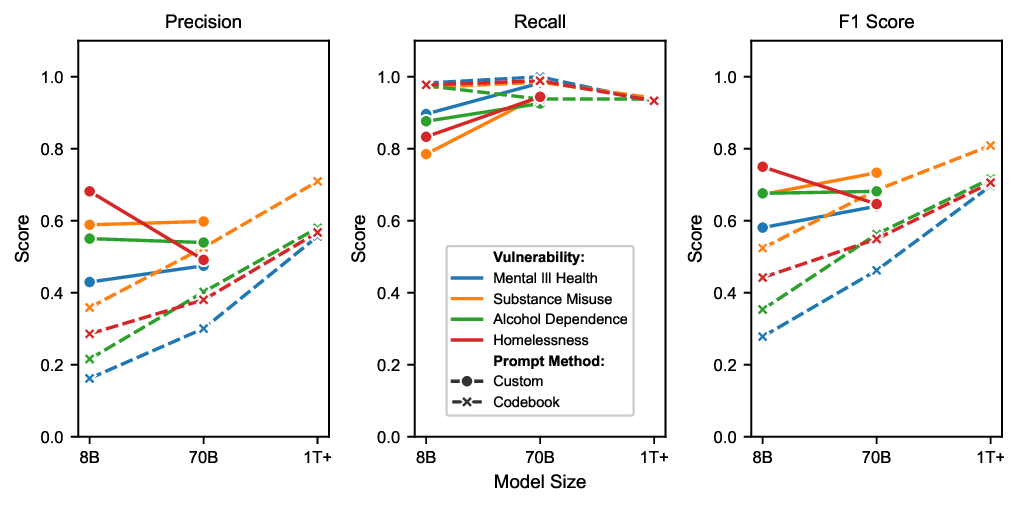}
    \caption{Precision, recall, and F1 scores for positive + inconclusive labels across different model sizes (8B, 70B, and 1T+) and prompt methods (Custom and Codebook). Solid lines with circles represent Custom prompts, while dashed lines with crosses represent Codebook prompts}
    \label{fig:prf1-plot}
\end{figure}

Fig. \ref{fig:prf1-plot} illustrates the precision, recall, and F1 scores for the combined positive labels across each LLM size and prompting method. The results align with trends observed in the MSE analyses. Codebook configurations exhibit a clear upward trend in performance as model size increases, primarily driven by improved precision. This is evident in the consistent increase in F1 scores across all vulnerabilities, with mental ill health and alcohol dependence showing particularly marked improvements from below 0.4 for 8B models to above 0.6 for 1T+ models. Custom prompt configurations demonstrate substantially enhanced performance compared to their respective codebook variants, especially for smaller models. The 8B custom prompt models achieve F1 scores (approximately 0.6-0.7) comparable to those of 1T+ codebook prompt models for most vulnerabilities.
All configurations demonstrate consistently high recall statistics, clustering above 0.8 and frequently exceeding 0.9. While the overall performance of custom configurations shows no clear relationship with model size, a closer examination reveals a consistent upward trend in recall as models get larger - the corresponding precision scores vary considerably between vulnerabilities, leading to fluctuating F1 scores that partially mask this recall improvement. However, in contexts where positive examples are relatively rare, high recall may be more valuable than balanced performance. These results suggest a promising practical application: the models could serve as effective screening tools, reliably identifying negative examples that can be excluded from manual review. This would allow human coders to focus their limited resources on examining only those cases flagged as positive or inconclusive by the model, potentially offering significant efficiency gains even if precision remains imperfect.

\subsubsection{Confusion Matrices}\label{subsubsec313}

\begin{figure}[htbp]
    \centering
    \includegraphics[width=0.98\textwidth]{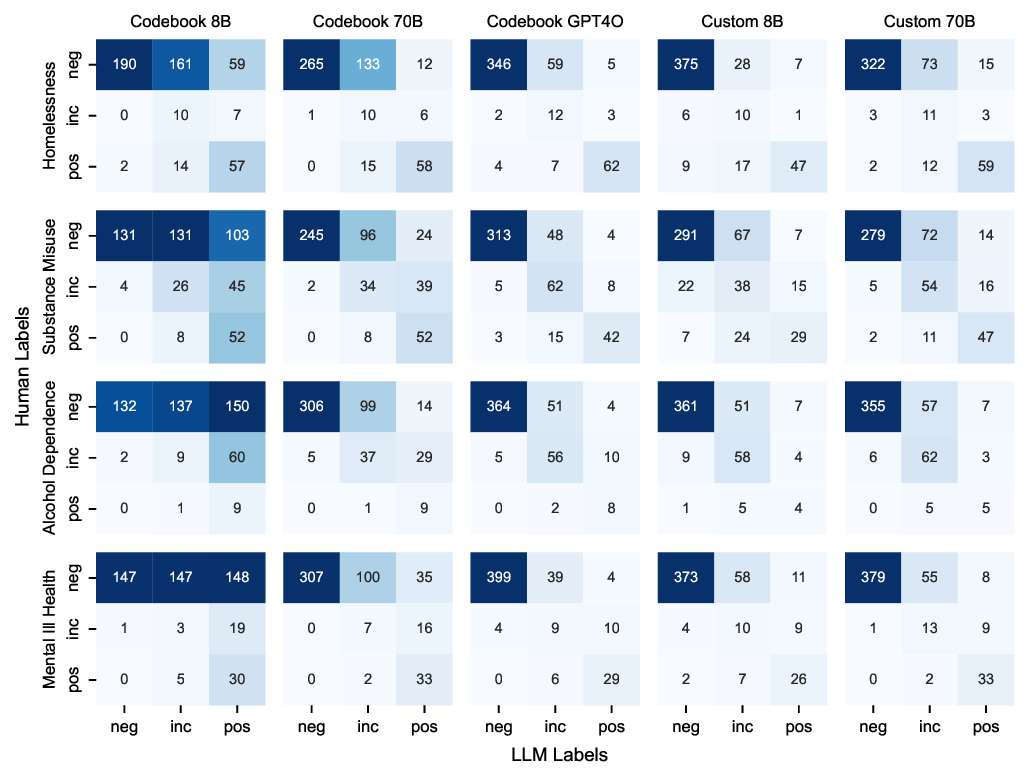}
    \caption{Confusion matrices comparing human labels (rows) with LLM consensus labels (columns) across different labelling configurations and vulnerability types. Cell values and shading intensity indicate the number of examples assigned each label combination. Darker shading indicates higher frequencies, with diagonal elements representing agreement between human and LLM labels. Results show strong alignment on negative classifications across all configurations, with most disagreements occurring at the boundaries between negative-inconclusive and inconclusive-positive categorizations}
    \label{fig:conf-mat}
\end{figure}

In order to further visualize the alignment between human and LLM labels, Fig. \ref{fig:conf-mat} shows confusion matrices for each labelling-configuration compared to the human labels. In each matrix, rows represent the human labels, columns represent the LLM labels, with the numbers in each cell (and the color of that cell) representing the number of examples assigned the respective labels. Within each square of nine cells for each labelling configuration, the cells along the diagonal from top-left to bottom-right represent agreement between the LLM and human labels, the off diagonals represent disagreement.

The confusion matrix analysis reveals additional nuances in model performance beyond those identified in the precision, recall, and F1 score metrics. Firstly, while all configurations demonstrate strong alignment with human coders on negative classifications, the analysis shows that disagreements primarily result from the LLMs over-assigning inconclusive labels to cases human coders judged as negative. This trend is particularly evident in the GPT-4o model, which generally aligns well with human assessments aside from a tendency to classify some human-negative cases as inconclusive. To illustrate, in Substance Misuse x GPT-4o, 48 cases categorized by humans as negative were categorized by the model as inconclusive. A secondary pattern emerges with the custom prompt configurations where, unlike GPT-4o, they show more variability at the inconclusive-positive boundary, assigning positive labels where human coders were more conservative with an inconclusive classification or vice versa. This is most prominent with the 70B model, suggesting that while the larger model demonstrates a greater precision in negative classifications, it also introduces a greater degree of variability in cases deemed inconclusive or positive.

We have already discussed how the strong alignment with negative human labels suggest LLMs may be effective as initial filters for excluding clearly negative cases. However, these results also suggest there might be further strategies for reducing manual labelling requirements by focusing review efforts where model-human disagreements most frequently arise. For instance, with GPT-4o, a strategy of targeting only the inconclusive labels for manual inspection could yield high overall accuracy, as its remaining classifications tend to align closely with human labels. A similar approach could be applied to the custom prompt configurations, although reviewing both inconclusive and positive cases may be advisable to refine estimates further, particularly where these models introduce variability between positive and inconclusive judgments.

\subsection{Label Variability}\label{subsec32}

The consensus labels analyzed thus far represent only a summary of each model’s output. However, every example underwent 10 labelling iterations per vulnerability and labelling configuration. The following analyses explore the consistency and variability within these multiple label assignments.

\subsubsection{Label Entropy}\label{subsubsec321}

To quantify the consistency of model outputs, we calculated the entropy of labels across repeated classifications. Entropy, in this context, measures the uncertainty in the labels assigned by each model by analyzing variability in the ten classifications (positive, inconclusive, negative) generated for each example by a given model. A higher entropy value indicates greater uncertainty or disagreement among the labels, while a lower entropy value suggests more consistent labelling. The entropy for a given narrative is calculated using the Shannon entropy formula in Eq.~(\ref{eq:entropy}):

\begin{equation}
    H = -\sum_{i=1}^{n} p_i\log_2(p_i)
    \label{eq:entropy}
\end{equation}

where $p_i$ is the probability of label $i$, and $n$ is the number of possible labels (in this case, 3). 

To estimate the average entropy and associated uncertainty for each labelling configuration we employed a bootstrapping approach. We resampled the entropy values with replacement 10,000 times, calculating the mean for each resample, and deriving the overall mean and 95\% confidence intervals from the distribution of these resampled means.

\begin{figure}[htbp]
    \centering
    \includegraphics[width=0.98\textwidth]{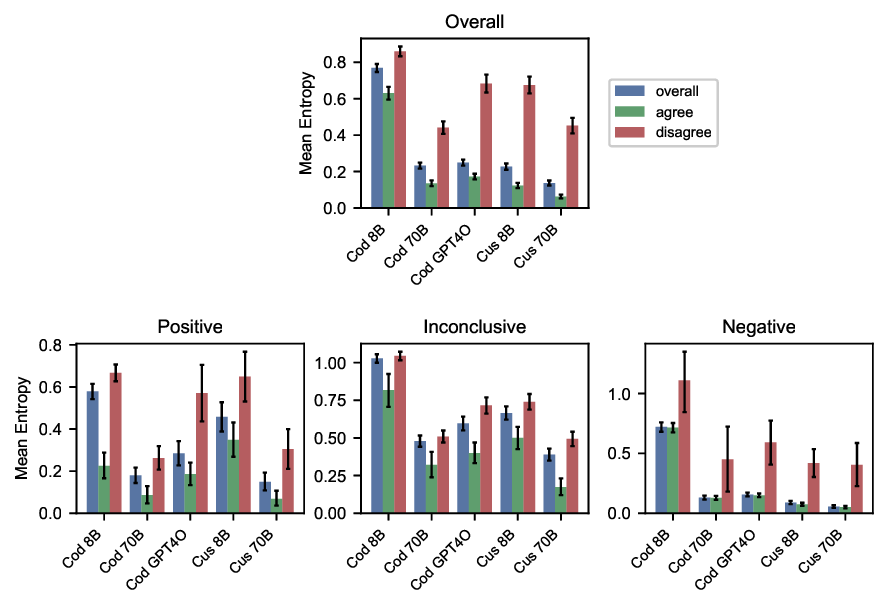}
    \caption{Mean entropy of LLM labels across different labelling configurations, with 95\% confidence intervals derived from bootstrap resampling. The top panel shows overall entropy, while bottom panels show entropy stratified by consensus label type (Positive, Inconclusive, Negative). Within each panel, bars represent overall entropy (blue), entropy for examples where LLM and human labels agree (green), and entropy for examples where they disagree (red). Lower entropy values indicate more consistent labelling across repeated classifications, with clear patterns showing lower entropy when LLM and human labels agree}
    \label{fig:entropy-bars}
\end{figure}

The entropy analysis in Fig. \ref{fig:entropy-bars} reveals a consistent theme across different labelling configurations and outcome scenarios: as models assign labels that agree with human judgements, they also exhibit greater certainty in these decisions. This pattern manifests in several key observations.

Firstly, instances where the model consensus agrees with human labels consistently show lower entropy than instances of disagreement (demonstrated by the relative size of the green bars with respect to the red bars). This trend persists across all labelling configurations, and holds true when examining individual label categories (positive, negative, and inconclusive). This relationship between label entropy and human-model disagreement indicates that model uncertainty could be a useful tool for automatically identifying and prioritizing ambiguous cases for expert review, potentially improving overall labelling accuracy.

The relationship between model-human agreement and certainty is further reinforced by the trends observed across different model sizes and prompting strategies. Consistent with earlier analyses comparing model outputs to human labels, labels from larger models and custom prompts generally exhibit lower entropy overall (though, a notable exception to this trend is the small increase in entropy between the 70B parameter model and GPT-4o). Moreover, the differences in the entropy values between examples agreeing with and disagreeing with human labels increases with model size, and when moving from the codebook to custom prompts, in line with the trends already observed. These findings indicate that as models become more adept at producing labels that align with human judgements, they also become more consistent in their classifications.

The label-specific entropy trends provide further evidence of this theme. The trends observed in the overall entropy statistics persist when divided into the different labels. Negative labels, which previous results showed to have the highest agreement between models and human coders, exhibit the lowest entropy overall. Conversely, inconclusive labels show the highest entropy overall, reflecting an inherent uncertainty in cases that, by definition, cannot be definitively classified. It is important to note that this higher entropy for inconclusive labels is partially deterministic, as our consensus method assigns an inconclusive label by default when there is no clear majority.

\subsubsection{Model Consensus \& Human Alignment}\label{subsubsec322}

Previous analyses revealed that higher agreement among LLM-generated labels correlates with improved alignment to human codings. This suggests that model agreement could serve as a useful proxy for confidence in LLM classifications. To explore this further, we investigated the relationship between model consensus (the extent to which repeated classifications for the same narrative agree) and alignment with human labels. Additionally, we examined how much of the dataset achieves varying levels of agreement, providing insights into the potential of using consensus as a guide for selective human review.

For each narrative, we counted how frequently each label (positive, inconclusive, or negative) appeared across its ten classifications. We focused our analysis on cases where a single label was assigned 6 or more times, as this represents a clear majority that cannot be matched by the other labels combined. For example, if a model classified a narrative as `negative' in 7 out of 10 iterations, this would represent an agreement level of 7.

For each agreement level, we assessed two key aspects:

\begin{enumerate}
\item \textbf{Proportion of Data by Agreement Level}: The percentage of narratives achieving each agreement level, stratified by label (positive, inconclusive, negative), vulnerability and labelling configuration. This quantifies how much of the dataset falls into categories with stronger or weaker model consensus.
\item \textbf{Alignment with Human Labels}: The proportion of classifications at each agreement level that matched human labels. This provides insights into how increasing agreement influences alignment, highlighting whether higher model consensus consistently leads to more accurate classifications.
\end{enumerate}

\begin{figure}[htbp]
    \centering
    \includegraphics[width=0.97\textwidth]{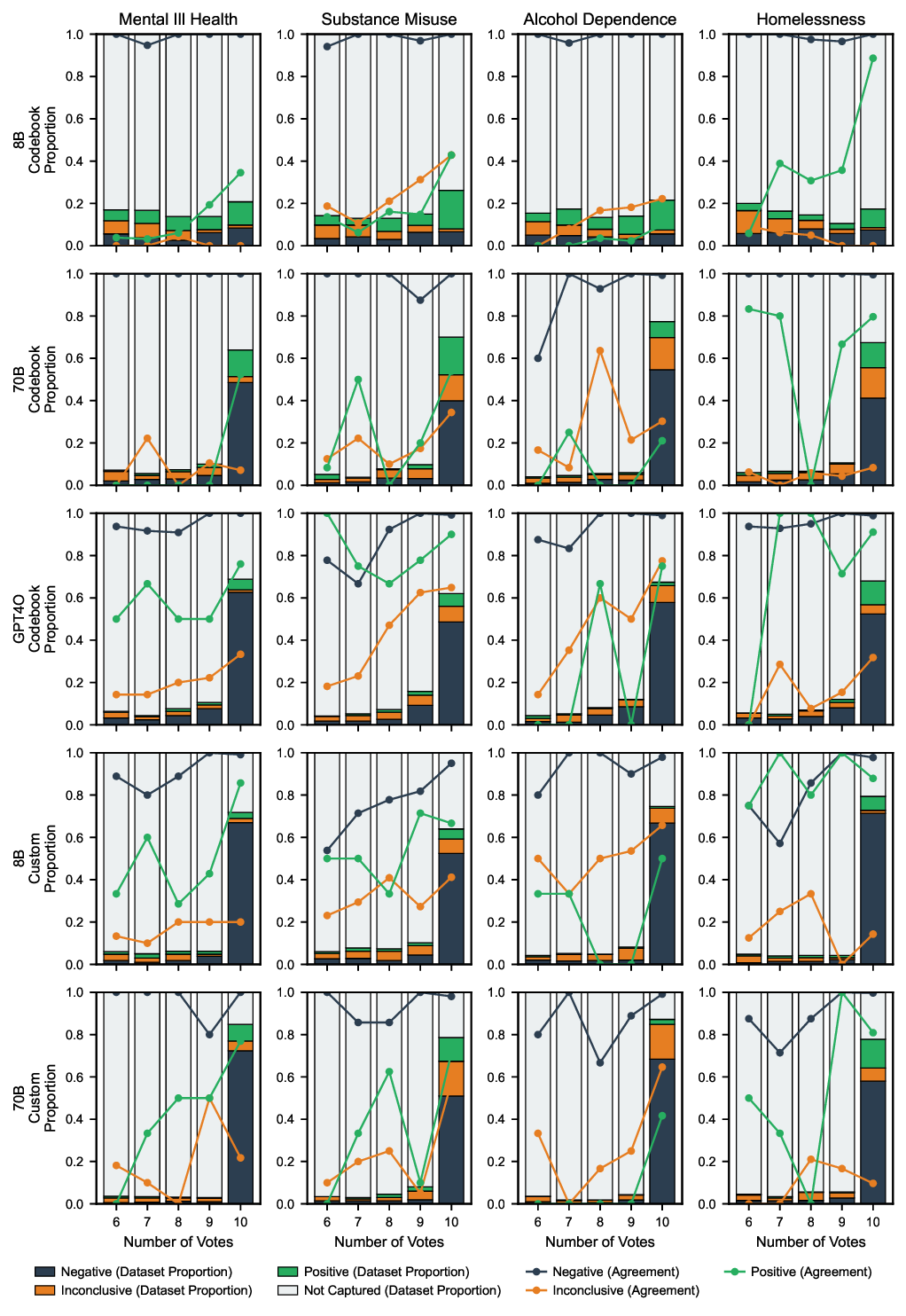}
    \caption{Relationship between model consensus and alignment with human labels across different labelling configurations and vulnerabilities. Stacked bars show the proportion of examples receiving 6-10 matching votes (x-axis) for each label type, with grey representing negative labels, tan representing inconclusive labels, and green representing positive labels. Line plots show the alignment between LLM and human labels at each consensus level for negative (black), inconclusive (orange), and positive (green) classifications. Higher consensus levels generally correspond to better alignment with human labels, particularly for negative classifications}
    \label{fig:votes}
\end{figure}

We visualized the results using stacked bar plots to display the proportion of narratives at each agreement level and label, with overlaid line graphs showing alignment with human labels, in Fig. \ref{fig:votes}.

Mirroring our entropy analyses, there is a strong relationship between the degree of consensus among model-generated labels and their alignment with the human labels, as shown by the rising trend in the line plots across all configurations - the more votes that agree on the same label, the more those classifications tend to align with human labels. With the exception of the 8B Codebook configuration, there is also a clear trend for the models to assign a majority of examples a unanimous 10 votes, signified by the dominating bars rightmost of each subplot. These patterns are particularly evident in negative classifications, which constitute the majority of high-consensus cases among the three best-performing configurations (GPT-4o, Custom 70B, and Custom 8B), ranging from 48-72\% of examples depending on the vulnerability and configuration, and aligning with human labels in greater than 95\% of cases at the 10/10 agreement level. For instance, GPT-4o's unanimous negative classifications align with human labels at rates of 100\% for mental ill health, 99.2\% for substance misuse, 99\% for alcohol dependence, and 98.9\% for homelessness at the highest agreement level. These results extend our previous findings on the effectiveness of LLMs as negative filters, demonstrating that near-perfect alignment with human coding can be achieved by flagging less confident classifications for review, while still maintaining automated classification for the majority of negative cases.

The results also indicate that the alignment of positive classifications might be improved by selecting examples at higher agreement levels, though with more variation across vulnerabilities and labelling configurations. As shown by the green sections of the bars and corresponding green lines, GPT-4o achieves the highest proportion and alignment for positive classifications at maximum agreement, though these represent a relatively small portion of the total examples overall. For example, at agreement level 10, GPT-4o identifies 11.2\% of homelessness cases as positive with 91\% alignment, and 6\% of substance misuse cases as positive with 90\% alignment. However, performance varies significantly by vulnerability and labelling configuration: substance misuse and alcohol dependence positive labels are typically not as well aligned as the other vulnerabilities (though, in the case of alcohol dependence this is largely related to the lower number of positive examples in the sample). The variable performance across different vulnerabilities suggests that while agreement levels could be used to improve the reliability of positive classifications, the appropriate threshold for automated versus manual coding would need to be carefully calibrated based on both the specific concept being coded and one's tolerance for potential misclassification.

Inconclusive classifications show distinct patterns from positive and negative labels in how they distribute across agreement levels. Unlike negative and positive classifications, which tend to concentrate at higher agreement levels (especially 9-10 votes), inconclusive labels show a notably flatter distribution across agreement levels. This pattern is consistent across all model sizes and prompting strategies and aligns with our conceptual understanding of what ``inconclusive'' represents - cases where evidence is mixed or ambiguous rather than clearly indicating presence or absence. These results suggest that LLMs are replicating this ambiguity in their repeated classifications, effectively ``disagreeing with themselves'' about whether cases are truly inconclusive or better classified as positive or negative. This is particularly evident in the alignment curves, where inconclusive classifications consistently show lower alignment with human labels even at high agreement levels. For instance, while GPT-4o's negative classifications reach near-perfect alignment at agreement level 10, its inconclusive classifications rarely exceed 70-80\% alignment even with maximum consensus. This pattern suggests that while LLMs can often effectively identify clear positive and negative cases, inconclusive classifications appear to warrant human review regardless of agreement level, reflecting the inherent complexity of cases where evidence is ambiguous or conflicting.

\subsection{Qualitative Insights}\label{subsec33}

Both of our prompting approaches instructed the models to provide chain-of-thought (CoT) explanations alongside their classifications, with the primary aim of anchoring the outputs to specific content from the narratives and the prompt instructions, thereby reducing the likelihood of fabrications. While these explanations might not necessarily reveal the true reasoning behind the models’ classifications, they provide a practical tool for interpreting outputs by linking features of the input narrative to the given instructions and the final label. This can help identify where models may diverge from instructions or where their outputs appear inconsistent with human-applied labels. In this context, the explanations offered a means of exploring potential patterns in model behavior, particularly in cases of disagreement. 

With this in mind, we conducted a short qualitative analysis of cases where the custom 8B parameter configuration’s classifications for substance misuse disagreed with human labels. This particular labelling configuration and vulnerability were selected for review because, among the higher-performing models with potential practical applications, it demonstrated the greatest divergence from human labels, offering a greater number and potential variety of disagreements to explore. Analyses revealed patterns of misclassification, such as a tendency to misattribute “unusual” behaviors— nervousness or erratic actions—as evidence of substance misuse, even in the absence of clear associations or instructions emphasizing these behaviors. Additionally, the model sometimes demonstrated inconsistent adherence to specific instructions; for instance, while prompts explicitly excluded alcohol-related evidence as an indicator of substance misuse, the model would sometimes follow this instruction, but on other occasions would cite alcohol consumption as justification for a positive or inconclusive label. At the same time, the model consistently followed other more complex, nuanced instructions, indicating that its errors were not the result of a general inability to align with detailed or complicated instructions. Unfortunately, the limited scope of our analyses prevents us from generalizing these findings or identifying their underlying causes. Addressing these issues in depth would require a more systematic approach, with additional experimental data exploring model behavior across configurations, a greater number of human labelers and written explanations for each of the human labels.

\subsection{Counterfactual Analyses}\label{subsec34}

To investigate potential biases in the language models and our methodology, we developed a counterfactual approach to assess the impact of key demographic characteristics on vulnerability classifications. This approach explores these potential biases by comparing model classifications for narratives where the sex or race descriptors of individuals are systematically manipulated.

\subsubsection{Counterfactual Method}\label{subsubsec341}

To generate data for these analyses we selected a subset of 100 narratives from our original dataset, ensuring approximately equal representation of the four vulnerabilities and the presence of a single, clearly identifiable subject. These narratives were manually annotated with the race and sex of the subject as described in the original text. Using a custom script employing GPT-4o, we generated counterfactual versions of each narrative, systematically altering the race and sex descriptors across a predefined set of demographics (sex: unknown, female, male; race: unknown, Black, White, Hispanic, Asian). This process yielded a set of 1500 counterfactual narratives (100 narratives, with 15 different combinations of sex and race) identical in content to the originals, differing only in the demographic descriptors of the subject.

We then applied our original classification methodology to this new dataset of counterfactual narratives, using both codebook and custom prompts with Llama 8B, Llama 70B, and GPT-4o models to classify each narrative for the four vulnerabilities of interest. The resulting dataset allows us to examine the potential impact of race and sex on the models’ vulnerability classifications, thereby assessing any systematic biases in our approach.

We employed generalized linear mixed-effects models (GLMMs) to assess the impact of demographic characteristics on vulnerability classifications. Analyses were conducted using R \citep{r_core_team_r_2024} with the lme4 package \citep{bates_fitting_2015}. For each combination of vulnerability, prompt type, and model, we fitted a GLMM with a binomial distribution and logit link function. The dependent variable was a binary outcome combining positive and inconclusive classifications (1) versus negative classifications (0). The model was specified as shown in Eq.~(\ref{eq:glmm}):

\begin{equation}
\label{eq:glmm}
\log\left(\frac{p_{ij}}{1-p_{ij}}\right) = \beta_0 + \beta_1\text{Race}_{ij} + \beta_2\text{Sex}_{ij} + u_j
\end{equation}

where $p_{ij}$ is the probability of a positive/inconclusive classification for observation $i$ in narrative $j$, $\beta_0$ is the intercept, $\beta_1$ and $\beta_2$ are the fixed effects for race and sex respectively (with `unknown' as the reference category), and $u_j$ is the random intercept for each base narrative.

We calculated average marginal effects (AMEs) for each demographic characteristic using the margins package \citep{leeper_margins_2024}. AMEs represent the average change in the probability of a positive/inconclusive classification associated with each demographic category, relative to the ‘unknown’ reference category. For each marginal effect, we computed point estimates, standard errors, 95\% confidence intervals, z-values, and p-values.

To address multiple comparisons, we applied the Holm-Bonferroni method to adjust p-values across all models and demographic characteristics tested. Effects with adjusted p-values \textless 0.05 were considered statistically significant, indicating a reliable association between the demographic characteristic and the probability of a positive/inconclusive vulnerability classification.

\subsubsection{Counterfactual Results}\label{subsubsec342}

\begin{figure}[htbp]
    \centering
    \includegraphics[width=0.98\textwidth]{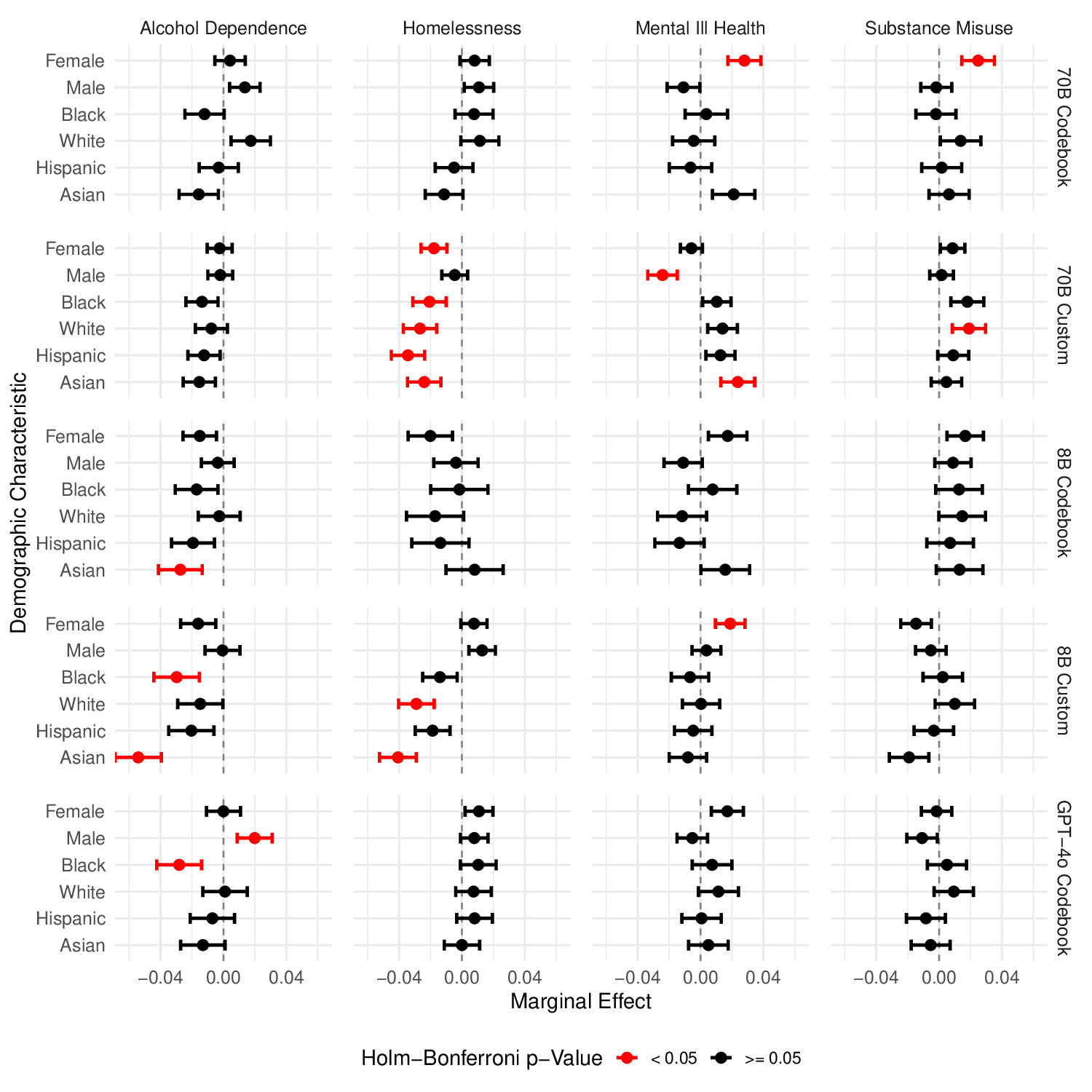}
    \caption{Average marginal effects of demographic characteristics (sex and race) on the probability of positive/inconclusive vulnerability classifications across different labelling configurations. Points show effect estimates relative to 'unknown' baseline categories, with horizontal lines indicating 95\% confidence intervals. Red points indicate effects that remain statistically significant after Holm-Bonferroni correction (p \textless 0.05), while black points indicate non-significant effects. Most demographic effects are small and non-significant, with few consistent patterns across vulnerabilities or labelling configurations}
    \label{fig:counterfac}
\end{figure}

Fig. \ref{fig:counterfac} shows the average marginal effects and confidence intervals of the demographic features for each combination of labelling approach and vulnerability. There are no consistent patterns across labelling approaches or vulnerability types and, after applying the Holm-Bonferroni correction for multiple comparisons, few statistically significant effects remain. The magnitudes of these effects are generally small, with the largest observed change in probability being 5.4\% (for the Asian race category in the 8B custom alcohol dependence model) - this is likely negligible in the context of the variability already observed in individual label assignments.

Notable findings include a set of significant race effects for homelessness classification using the 70B custom model. All specified race categories showed a reduced likelihood of positive labelling compared to the ‘unknown’ race baseline, with effect sizes ranging from -2.07\% to -3.44\%. This suggests that rather than indicating differential treatment among specified races, narratives containing individuals of unspecified race are more likely to be flagged as positive for homelessness. Similar patterns, albeit with more dispersed effects, were observed for the 8B custom model in both homelessness and alcohol dependence classifications.
The largest model, GPT-4o, showed few significant demographic effects across vulnerabilities. However, in alcohol dependence classification, being black was associated with a 2.81\% lower probability of a positive label and being male with a 1.99\% higher probability. Unlike other models, which exhibited more varied effects across demographics, GPT-4o consistently showed little to no influence of demographic factors in most cases. This consistency makes the specific effects in alcohol dependence classification particularly noteworthy and suggests a potential area for further investigation, especially given GPT-4o’s superior overall performance.

\section{Discussion}\label{4}

Our results demonstrate that instruction-tuned large language models can effectively support qualitative coding of police narratives, particularly in identifying and filtering cases where vulnerabilities are absent. This capability, combined with some strong performance in identifying clear positive cases with high confidence, and limited evidence of any demographic bias, suggests promising applications in augmenting traditional qualitative analysis approaches.

Our analysis reveals important patterns in how model size and prompting strategy affect performance. While larger models demonstrated incrementally better performance with standard codebook instructions, custom prompts significantly improved the performance of smaller models with respect to their codebook counterparts. The 8-billion parameter model with custom prompts achieved F1 scores that were comparable or even superior to GPT-4o using codebook instructions (Mental Health: 0.58 vs 0.70; Homelessness: 0.75 vs 0.71; Substance Misuse: 0.67 vs 0.81; Alcohol Dependence: 0.68 vs 0.72). This finding has significant practical implications. Larger models offer clear advantages - they can effectively utilize detailed codebook instructions without extensive prompt engineering, generally achieve better performance, and may be the optimal choice when resources permit their use. However, the strong performance of smaller models with custom prompts demonstrates their viability as alternatives, particularly valuable in contexts where data security concerns preclude sharing information with proprietary LLM providers.

The models' effectiveness as negative filters emerge as one of our most promising findings. Even using simple consensus labels, all configurations demonstrated strong capabilities in identifying narratives without vulnerability indicators, with precision for negative classifications consistently exceeding 90\% across vulnerabilities. This performance improves further when considering label agreement levels. At maximum agreement (10/10 classifications), the three best-performing configurations achieved remarkable precision: Custom 8B showed alignment rates of 95-99\% while classifying 52-71\% of examples as negative, Custom 70B achieved 98-100\% alignment on 51-72\% of examples, and GPT-4o reached 99-100\% alignment on 49-63\% of cases. These results suggest that LLMs could dramatically reduce manual coding requirements by automatically filtering out clear negative cases, allowing human coders to focus their efforts on potentially positive cases. The consistency of these results across model sizes and prompting strategies is particularly encouraging, suggesting that effective negative screening could be implemented even in contexts where smaller models are preferred for practical or security reasons.

Analysis of positive and inconclusive classifications indicates substantially lower reliability when compared with negative labels. Though performance generally improves with higher levels of agreement between repeated classifications, even under optimal conditions accuracy remains limited. The strongest results were observed with GPT-4o - for unanimously classified positive cases, it achieves precision exceeding 90\% for both substance misuse and homelessness, though these highly confident positive classifications represent only a small portion of actual positive cases (6-11\% of examples). However, model performance varies substantially across vulnerability types and becomes considerably less reliable for inconclusive classifications. Even at high agreement levels, precision for inconclusive labels rarely exceeds 70\%, with considerable variation across vulnerabilities and models. This pattern aligns with confusion matrix analyses showing that models tend to err toward inconclusive labels when uncertain, more frequently marking true positives as inconclusive than negative, and true negatives as inconclusive rather than positive. These findings suggest that while some positive classifications might be automated depending on one's tolerance for error, inconclusive labels appear to function more as indicators of model uncertainty than as meaningful classifications, likely warranting human review in most cases. It's worth noting that these patterns emerge in a dataset where negative cases predominate (approximately 85-90\% of examples), and further research with more balanced label distributions would be valuable to fully understand the relative strengths and limitations of these models across different contexts.

The counterfactual analyses provide broad reassurance regarding demographic biases explored, while highlighting the importance of careful monitoring in new applications. After correcting for multiple comparisons, we found remarkably few statistically significant demographic effects across our configurations, and where present, their magnitudes were generally small (\textless 5\% change in classification probability). The largest model, GPT-4o, showed particularly encouraging results, with near-zero demographic effects for most vulnerabilities. The 8B and 70B custom prompted models showed several statistically significant differences in classification rates between demographic groups, but these effects mostly suggested a tendency to label more examples as positive when demographic information isn't specified, rather than indicating differential treatment between racial groups. While these effects are minor relative to the overall classification variability we observed, they emphasize the importance of ongoing monitoring for potential biases, particularly when deploying these systems at scale where small effects could accumulate into meaningful disparities.

\subsection{Practical Implications: Benefits \& Limitations}\label{subsec41}

The application of IT-LLMs to qualitative coding offers several compelling advantages while raising important considerations for implementation. Perhaps most significantly, these models enable analysis at scales impractical for traditional qualitative methods. While manual coding of thousands of narratives typically requires weeks or months of sustained effort, IT-LLMs can process comparable volumes of text rapidly, enabling more comprehensive analyses of routinely collected data that have historically been constrained by resource limitations.

IT-LLM workflows also provide additional ways to maintain or enhance methodological rigor in qualitative research. The process of developing prompts for IT-LLMs inherently requires researchers to fully articulate their classification criteria through explicit codebooks and formal prompt engineering, making analytical decisions more transparent and replicable. While human coders may unconsciously supplement written definitions with implicit knowledge or reach shared understandings through discussion, IT-LLMs work solely from the explicit instructions provided. Moreover, qualitative researchers often employ resource-intensive validation steps like intercoder reliability testing to assess coding credibility \citep{oconnor_intercoder_2020}. The non-deterministic nature of IT-LLMs offers an alternative approach: researchers can generate repeated classifications for all examples, including appropriate counterfactuals as we explore here, to formally assess model consistency. While LLMs may still exhibit biases, these stem from training data and prompt design rather than the moment-to-moment subjective judgments that characterize human coding.

However, several important limitations warrant consideration. Our qualitative analyses revealed that models do not always strictly adhere to provided instructions, sometimes over-interpreting behaviors as vulnerability indicators despite explicit guidance to the contrary. While careful prompt engineering can minimize such issues, they highlight an ongoing need for human oversight, particularly in ambiguous cases. More fundamentally, LLMs lack true transparency in their decision-making processes. Though they provide natural language explanations for classifications, these represent post-hoc rationalizations rather than insights into how inputs map to outputs or why repeated classifications may vary. This opacity complicates validation efforts and may limit their application in more interpretive qualitative analyses that seek to understand underlying meanings or contexts.

Practical implementation also raises important technical considerations. Large proprietary models like GPT-4o, while highly capable, require sharing data with third-party servers - potentially problematic for sensitive research data. Smaller open-source models deployed locally offer greater security but demand more technical expertise and computational resources. Furthermore, proprietary models may be modified by developers without notice, potentially altering performance or introducing new biases. This instability necessitates ongoing validation to ensure models remain fit for purpose, particularly for longitudinal research projects.

These limitations suggest IT-LLMs are best viewed as tools to augment rather than replace traditional qualitative methods. Their ability to rapidly process large volumes of text while maintaining consistent criteria makes them valuable for initial screening and filtering tasks. However, the need for human oversight of ambiguous cases, combined with challenges around transparency and stability, indicates they should complement rather than supersede expert judgement. Used thoughtfully within these constraints, IT-LLMs offer promising capabilities for expanding the scope and of qualitative research methodology.

\subsection{Limitations \& Recommendations for Future Work}\label{subsec42}

\subsubsection{Label Ambiguity}\label{subsubsec421}

One of the primary challenges in our study is the inherent ambiguity in identifying vulnerabilities within FIO narratives. The narratives analyzed in this study were not written with a focus on capturing specific vulnerabilities like mental ill health or homelessness, so identifying vulnerability often requires interpreting indirect cues. This lack of a clear ground truth introduces substantial subjectivity, as both human and model labelling hinge on whether the individual “sounds like” they are vulnerable. Vulnerabilities often exist on a continuum, and setting strict boundaries between categories—such as “definitely present” or “definitely absent”—is arbitrary. Consequently, alignment between model and human labels doesn’t always confirm correctness; it may simply reflect shared interpretations of ambiguous information.

In this study, this ambiguity was compounded by our limited labelling resources, with only two human coders and a third reviewer for adjudicating disagreements. This limited pool forced us into a binary “agree/disagree” metric rather than a more nuanced evaluation of model performance. A larger group of human coders could better capture the full range of interpretations, providing a more comprehensive benchmark. For instance, areas of high human disagreement could indicate cases where the LLMs’ alternative interpretations are equally valid. Such an approach could reveal that model disagreement with human coders is, in part, reflective of ambiguity inherent to the data, rather than model error. We recommend that future work incorporates more human labels to better capture inter-rater variability, using this broader consensus to refine assessments of model performance and clarify the interpretive challenges posed by ambiguous data.

\subsubsection{Custom Prompt Development}\label{subsubsec422}

Another limitation of our study involves the lack of systematic guidance on prompt engineering. While we developed custom prompts for IT-LLMs, the iterative prompt development process was conducted outside the documented experimental results, with adjustments and refinements not formally included in our reported methods. Prompt engineering is inherently iterative and interpretive, as minor changes in wording, format, or even syntax can lead to significant variations in model outputs. However, our study does not explore how such variations affect performance, limiting the replicability of our approach for other researchers. 

Further, we do not identify specific elements of our prompts that most contributed to accuracy, nor do we provide generalizable methods for adapting prompts to different datasets or coding tasks. Without a systematic evaluation of prompt variations, it is difficult to determine the extent to which our results generalize to other domains. We recommend that future work systematically examine prompt development to identify best practices, which could yield clearer guidelines for structuring prompts effectively across varied domains and tasks.

\subsubsection{Practical Limitations}\label{subsubsec423}

This study focuses on applying IT-LLMs solely for deductive coding in research settings, with no assessment of their potential use in practical policing settings. While it is not difficult to envisage more applied use-cases, the relative novelty of evaluations such as this dictate that our approach is purposefully limited to exploring aggregate trends within a large dataset and does not extend to case-by-case analysis. While we believe that there is considerable promise in applying IT-LLMs to derive aggregate-level insights, the approach still lacks the consistency and accuracy required for individual case assessments, where the stakes are inherently higher, and outcomes hinge on precise judgments. The non-deterministic nature of LLMs and their occasional inconsistencies underscore the importance of confining this method to such exploratory analyses rather than operational decision-making contexts.

Further, while aggregate analyses like these may contribute to a general understanding of vulnerability markers across datasets, we have not examined any practical implications for decision-making based on these findings. Caution is essential when interpreting our results; we strongly advise against using models of this type to make decisions on individual or case-specific bases, nor can we endorse operational applications based solely on aggregate trends. Instead, we propose that IT-LLM derived insights support one element of a broader collection of focused problem-based analyses, enabling insights to be derived from unstructured data previously inaccessible due to resource constraints. Future work might explore the potential for practical applications, but we emphasize that any such use would require meticulous, ongoing and objective evaluation of the method’s reliability, ethical implications, and alignment with fair and transparent policing practices.

\subsubsection{Model Advancement}\label{subsubsec424}

As a final consideration, it is important to recognize that our evaluation represents a focused assessment of IT-LLMs on a specific task - deductive coding of vulnerability in police narratives - at a particular moment in the rapid evolution of these technologies. While our results demonstrate promising capabilities even with these early models, they likely represent a baseline rather than a ceiling for such applications. Indeed, the models' ability to follow complex instructions, reason about evidence, and explain their classifications suggests potential for more sophisticated applications where they act as collaborative partners throughout the qualitative coding process rather than serving purely as classification tools. Future work might explore workflows where models assist in codebook development by identifying potential ambiguities in definitions, suggest refinements based on patterns in their own uncertainty, or engage in more dynamic dialogue with research teams about challenging cases. Such applications could enhance qualitative analysis workflows in ways that thoughtfully combine human insight with machine-assisted analysis, though careful evaluation of reliability and validity would remain essential. While the present study focuses necessarily on basic classification capabilities, the models' demonstrated capacity for reasoned analysis suggests valuable directions for future research into more comprehensive applications of IT-LLMs in qualitative analysis. 

\newpage
\bibliography{zotero_bib}  

\newpage
\begin{appendices}

\section{Codebook Definitions}\label{secA}

\subsection{Mental Health Difficulties}\label{secA1}

\noindent\textbf{General Definition:}\medskip

This category should include reports of behaviors, statements, or circumstances that suggest an individual may be experiencing mental health challenges. This includes explicit evidence of mental health diagnoses, or specific behaviors related to mental health problems such as self-harm or suicidal behavior. It can also manifest as disorientation, irrational behavior, speaking incoherently, visible distress without clear cause, or descriptions of behavior that significantly deviate from social norms without an obvious immediate cause.

\medskip\noindent\textbf{Categories:}

\begin{enumerate}
    \item \textbf{Positive:} Clear indications of mental health challenges based on behavior, statements, or context within the report, and having ruled out mediating factors such as intoxication or situational stressors. Examples include:
    \begin{itemize}
        \item Explicit discussion of mental health problems, diagnosed mental health conditions or engagement with mental health services
        \item Indications of self-harm or suicidal behavior
        \item Incoherent or nonsensical speech (in the absence of other factors such as intoxication)
        \item Disorientation or confusion about surroundings, time, or reality more generally (in the absence of other factors such as intoxication/head injury/shock)
        \item Extreme irrational behavior or reactions that are clearly unrelated to situational factors
        \item References to hallucinations, delusions, or paranoia without any indication of drug use
        \item Evidence of a pattern of erratic or highly unusual behavior over multiple encounters/incidents
    \end{itemize}
    
    \item \textbf{Inconclusive:} Cases where behavior might suggest mental health difficulties, but there are possible alternative explanations, or there is insufficient evidence to determine if mental health issues are present. Examples include:
    \begin{itemize}
        \item Confused, erratic or delusional behavior that could be attributed to drugs/alcohol, but there isn't evidence to confirm either way
        \item Situations where individuals exhibit extreme/disproportionate emotions or unusual behavior, where there is insufficient detail to determine if circumstances justify the behavior (e.g. extreme disagreements, seemingly unprovoked hostility, incoherent statements)
    \end{itemize}
    
    \item \textbf{Negative:} Absence of any explicit indicators of mental health issues, or indicators that can be attributed to confirmed situational factors like intoxication, anger, or typical stress responses.
    \begin{itemize}
        \item Rational behavior and coherent communication
        \item Behavior clearly motivated or mediated by the circumstances e.g. intoxication or drug abuse, situational stressors, shock or head injury
        \item Unusual behavior resulting from engaging in criminal activity, interaction with the police, or attempts to conceal criminal activity from the police
    \end{itemize}
\end{enumerate}

\subsection{Drug Abuse}\label{secA3}

\noindent\textbf{General Definition:}\medskip

``Drug abuse'' refers to incidents where individuals are using substances known for their high potential for abuse and dependency – these may include opiates and opioids, the smoking of methamphetamine or crack cocaine, and exclude alcohol and narcotics not commonly thought to be drugs of abuse e.g. marijuana, MDMA, hallucinogens, and recreational use of cocaine. This includes behaviors or circumstances that suggest an individual is actively using these substances in a way that could be detrimental, even if negative health impacts or dependency are not detailed in the report.

\medskip\noindent\textbf{Categories:}

\begin{enumerate}
   \item \textbf{Positive:} Clear indications of drug abuse based on behavior, physical evidence, or context within the report. This focuses on drugs with a high abuse potential that are typically associated with dependency and significant negative health or social outcomes. Examples include:
   \begin{itemize}
       \item Observable signs of impairment, that are specifically related as relating to drug use (as distinct from alcohol or other causes). Examples include severe disorientation, drowsiness, inability to communicate coherently, profuse sweating, or physical symptoms suggesting recent high-dosage use (e.g., track marks, unconsciousness).
       \item Direct admissions of using drugs of abuse, particularly in a context suggesting habitual use or dependency.
       \item Finding an individual in possession of drugs or drug paraphernalia in a setting that strongly suggests they are actively using them. Examples include drug packaging, syringes, pipes, and spoons with residue.
       \item Emergency medical responses to overdoses of substances known for their abuse potential.
       \item Current or past involvement or engagement with drug treatment or rehabilitation organizations
   \end{itemize}
   
   \item \textbf{Inconclusive:} Cases where drug use is possible based on the context or evidence, but the information is not sufficient to conclusively determine active abuse of drugs. Examples include:
   \begin{itemize}
       \item Medical or police interventions where drug use is one of several possible explanations for the individual's condition or behavior.
       \item Reports of intoxication where it is unclear if the individual is under the influence of narcotics (high) or has been drinking alcohol (drunk).
       \item Context that suggests drug use, such as the presence of drug paraphernalia, where it is unclear if any of the individuals present are actively using drugs
       \item Information detailing the use of drugs, where it is unclear if the drugs in question are drugs of abuse or are narcotics that aren't associated with dependence (marijuana, MDMA, hallucinogens, recreational use of cocaine)
   \end{itemize}
   
   \item \textbf{Negative:} Absence of any indicators suggesting individuals are active drug abusers. The circumstances or behavior could be associated with the sale or distribution of drugs, or drinking alcohol, or the use of narcotics that aren't considered drugs of abuse. Examples include:
   \begin{itemize}
       \item Incidents involving alcohol or other substances not classified as highly abusive e.g. marijuana, MDMA, hallucinogens, recreational use of cocaine
       \item Any circumstances or behavior associated with the sale or distribution of drugs or drug paraphernalia, in the absence of any evidence for active use
       \item Discussion of drug offences or convictions, where it isn't clear that the offences relate to drug use (as opposed to supply/distribution), or that the individual is currently using drugs
   \end{itemize}
\end{enumerate}

\subsection{Alcohol dependence}\label{secA2}

\noindent\textbf{General Definition:}\medskip

Encompasses behaviors, statements, or circumstances that suggest an individual may be experiencing alcohol dependence or may be experiencing challenges related to excessive alcohol consumption, such as severe intoxication that notably affects their behavior during an interaction. This includes clear signs of current intoxication, references to habitual heavy drinking, and contexts that strongly imply ongoing alcohol abuse.

\medskip\noindent\textbf{Categories:}

\begin{enumerate}
    \item \textbf{Positive:} Clear evidence of alcoholism or severe and problematic intoxication excluding that associated with social drinking, based on behavior, statements, or context within the report, sufficiently distinguished from other influencing factors such as drug use or transient emotional distress. Examples include:
    \begin{itemize}
        \item Specific evidence of frequent and heavy alcohol use (as distinct from narcotics) that suggest dependency or habitual misuse
        \item Engagement with or the recommendation of alcohol dependence services, such as AA or rehabilitation centers, where it is clear that the individual is an active drinker or is not in remission
        \item Disruptive, aggressive or problematic behavior explicitly attributed to alcohol consumption, excluding any settings associated with social drinking (bars/parties/festivals etc.)
        \item Repeated incidents or a known history of alcohol-related interactions with law enforcement, emergency services or alcohol-related services.
    \end{itemize}
    
    \item \textbf{Inconclusive:} Cases where there are potential indications of alcohol-related issues, excluding those associated with social drinking, but there is a lack of sufficient information to determine if alcohol is a problem or the extent to which alcohol is a problem. Examples include:
    \begin{itemize}
        \item Signs of heavy intoxication, excluding social drinking, where it is unclear if the individual is under the influence of alcohol or other drugs, or has another impairment.
        \item Signs that an individual is intoxicated, excluding social drinking, but it is questionable whether the intoxication is a sign of dependence, is causing any problematic behavior, or more generally is a reason for concern.
        \item Incidents where an individual's problematic behavior may be influenced by alcohol, excluding social drinking, but it is unclear the extent to which alcohol has contributed, or there is also significant potential for other contributing factors like mental health issues or situational stressors.
    \end{itemize}
    
    \item \textbf{Negative:} No significant indicators of excessive drinking or alcohol abuse are present, or excessive drinking in a social setting. Examples include:
    \begin{itemize}
        \item Any discussion of social drinking, including disruptive, problematic or aggressive behavior associated with having drunk excessively in a social setting
        \item Discussion of alcohol or evidence of drinking where there isn't reason to suspect problem/excessive drinking
    \end{itemize}
\end{enumerate}

\subsection{Homelessness}\label{secA4}

\noindent\textbf{General Definition:}\medskip

Situations where individuals lack a stable, permanent, and adequate nighttime residence. This includes both explicit declarations of homelessness and implicit indicators observable through behavior, circumstances, or environmental context reported. The definition aims to identify individuals who do not have access to consistent and private nighttime accommodations, which could include those temporarily staying in shelters, cars, or other non-residential settings.

\medskip\noindent\textbf{Categories:}

\begin{enumerate}
   \item \textbf{Positive:} Clear evidence that an individual is experiencing homelessness, either through self-report or unmistakable circumstances indicating a lack of stable housing. Examples include:
   \begin{itemize}
       \item Individuals explicitly stating they are homeless or do not have a home.
       \item Possession of a large number of personal belongings in public spaces, indicative of no permanent residence.
       \item Possession of sleeping equipment such as tents, sleeping bags or pillows in public
       \item Clear use of public spaces, squats or makeshift shelters as a primary sleeping arrangement
       \item Interactions with social or homelessness services that explicitly indicate their homeless status
   \end{itemize}
   
   \item \textbf{Inconclusive:} Cases where there are signs that may suggest homelessness, but there are possible alternative explanations, or there is insufficient evidence to definitively categories the individual as homeless. Examples include:
   \begin{itemize}
       \item Presence in settings commonly associated with homelessness, such as public spaces frequented by the homeless, abandoned buildings, or squats where it is unclear if the individual has stable housing
       \item Evidence of temporary or unstable residential accommodation as a primary nighttime residence e.g. hotels, AirBnBs, half-way houses, where it is suggested an individual doesn't have a more stable alternative
       \item being found asleep in public where it is unclear if the individual may have stable housing e.g. passing out when drunk, sleeping in a car
       \item Appearing very disheveled or unkempt in a manner that suggests a lack of access to bathroom facilities
   \end{itemize}
   
   \item \textbf{Negative:} Absence of any direct evidence for homelessness. The individual's circumstances or behavior can be clearly attributed to other factors that don't relate to a lack of stable housing. Examples include:
   \begin{itemize}
       \item Individuals whose behavior is linked to other vulnerabilities such as mental health issues or intoxication, with no other signs of housing instability.
       \item loitering or causing a nuisance in public spaces without other evidence of homelessness
       \item shoplifting, begging or other criminal activity associated with, but not evidence for, homelessness
   \end{itemize}
\end{enumerate}

\section{Custom Prompts}\label{secB}

\subsection{Custom prompt template:}\label{secB1}

\begin{verbatim}
    You are required to classify police incident reports for involvement of 
    persons experiencing {{ vulnerability }}. Use the following definitions for 
    the labels you should assign:

    POSITIVE: Report confirms that someone is experiencing {{ vulnerability }}, 
    or contains unmistakable evidence of {{ vulnerability }} having ruled out 
    any other plausible explanations. For example:

    {{ positive_evidence }}

    INCONCLUSIVE: Report contains evidence that is best explained by an 
    individual experiencing {{ vulnerability }}, but there is not definitive or 
    conclusive confirmation of {{ vulnerability }}. For example:

    {{ inconclusive_evidence }}

    NEGATIVE: Evidence for {{ vulnerability }} that can be explained by other 
    factors, or no evidence for {{ vulnerability }}. The following should not 
    be considered evidence for {{ vulnerability }}:

    {{ negative_evidence }}

    Write short notes highlighting quotes from the report, and link each quote 
    to the relevant quote above. Keep the notes to two sentences max.

    End your report with a classification that aligns with the evidence you 
    have highlighted. Use the format "Classification: [POSITIVE, INCONCLUSIVE, 
    NEGATIVE]". If the final word of your report is not the classification, it 
    will be marked invalid.
\end{verbatim}

\subsection{Mental Ill Health Evidence:}\label{secB2}

\begin{verbatim}
    Positive Evidence:
    - Statements that someone is experiencing mental health difficulties, has 
    or is receiving treatment for a mental health diagnosis
    - Individuals engaging with or being offered mental health services
    - Clear and obvious descriptions of emotional disturbance, irrational or 
    erratic behavior, statements or actions suggesting delusions or loss of 
    contact with reality, where there is no other reasonable explanation other 
    than someone experiencing mental health issues
    - Deliberate self harm or attempts at suicide

    Inconclusive Evidence:
    - Clear and obvious symptoms or behavior that are best explained as mental 
    health related, but there are other possible explanations for the behavior 
    symptoms (e.g. intoxication, circumstances not detailed in the report)

    Negative Evidence:
    - any evidence associated with drug/alcohol use or dependence, overdose or 
    detox/rehab services
    - any evidence associated with homelessness or street outreach activity
    - general stress, anger, aggression, hostility, panic or anxiety
    - strange or unusual behavior related to criminal activity or attempts to 
    conceal criminal activity
    - Anxiety in the presence of the police, or strange, erratic or aggressive 
    behavior towards the police
    - Evidence of medical treatment or medicines that are not confirmed to be 
    mental health related
\end{verbatim}

\subsection{Substance Misuse Evidence:}\label{secB3}

\begin{verbatim}
    Positive Evidence:
    - Explicit statements that someone is currently using drugs or is under the 
    influence of drugs
    - Individuals engaging with or being offered drug rehabilitation services 
    or organizations

    Inconclusive Evidence:
    - possession or presence of drugs or drug paraphernalia, with strong 
    evidence that the individuals present are actively using drugs
    - Discussion of rehabilitation services for an active addiction (excluding 
    general street outreach services), that might be drug or alcohol related 
    but it isn't specified 
    - Statements that an individual is intoxicated where it is unclear if the 
    intoxicant is alcohol or other narcotics

    Negative Evidence:
    - Any drug related activity, possession, sale or trafficking that isn't 
    accompanied by specific evidence that an individual is using drugs or has a 
    drug abuse problem
    - Marijuana use or possession
    - Any evidence associated with alcohol, homelessness, or mental health 
    conditions
\end{verbatim}

\subsection{Alcohol Dependence Evidence:}\label{secB4}

\begin{verbatim}
    Positive Evidence:
    - Explicit statements that someone is an alcoholic, frequently drinks 
    heavily, or is dependent on alcohol
    - Discussion of alcoholism services, such as rehabilitation centers 
    (specifically for drinking) or AA
    - Disruptive, aggressive or problematic behavior that is explicitly 
    attributed to alcohol consumption, excluding settings associated with social 
    drinking (bars/parties/festivals etc.)

    Inconclusive Evidence:
    - Disruptive, aggressive or problematic behavior that is clearly attributed 
    to intoxication, but it is not clear if drugs or drugs or alcohol are the 
    intoxicant
    - Confirmation of extreme alcohol intoxication, where it isn't clear if it 
    is a recurrent problem
    - Instances where extremely problematic behavior co-occurs with alcohol 
    consumption, but the extent of the influence of alcohol is unclear

    Negative Evidence:
    - Disruptive, aggressive or problematic behavior that isn't clearly 
    attributed to intoxication
    - Evidence of drinking, including intoxication in social drinking settings 
    - bars/festivals/parties
    - The presence of alcohol or alcohol containers where there isn't a clear 
    reason to suspect problem drinking
    - Any evidence relating to drug abuse, overdose, or homelessness
\end{verbatim}

\subsection{Homelessness Evidence:}\label{secB5}

\begin{verbatim}
    Positive Evidence:
    - Statements that someone is homeless or does not have any nighttime 
    accommodation
    - Individuals engaging with or being offered homelessness services or 
    organizations
    - Individuals being found with makeshift sleeping/living arrangements on 
    the street or in unstable living environments

    Inconclusive Evidence:
    - Strong evidence that someone does not have any nighttime accommodation, 
    but is not definitive
    - Being found asleep in public

    Negative Evidence:
    - Causing a nuisance, loitering, or being trespassed from places where 
    homeless individuals may congregate 
    - Use of detox or rehab services for alcohol or drug abuse
    - Any evidence or behavior associated with a person's drug use, 
    alcoholism, mental health difficulties, or sex work
    - Any vague or uncooperative responses to police questioning about 
    address information that don't result in an admission of homelessness
\end{verbatim}

\end{appendices}

\end{document}